\documentclass[journal]{IEEEtran}

\ifCLASSINFOpdf
\else
   \usepackage[dvips]{graphicx}
\fi
\usepackage{url}

\usepackage{graphicx}
\usepackage{hyperref}
\usepackage{xcolor}
\usepackage[linesnumbered,ruled]{algorithm2e}
\usepackage{amsmath}
\usepackage{booktabs}  
\usepackage{multirow}
\usepackage{subfigure}

\usepackage[capitalize]{cleveref}
\crefname{figure}{Fig.}{Figs.}
\crefname{section}{Section}{Sections}
\crefname{table}{Table}{Tables}
\crefname{algorithm}{Algorithm}{Algorithms}
\crefname{equation}{Eq.}{Eqs.}

\begin{document}

\title{Spiking Layer-Adaptive Magnitude-based Pruning}

\author{Junqiao Wang, Zhehang Ye, Yuqi Ouyang
\thanks{Junqiao Wang and Zhehang Ye are with Sichuan University-Pittsburgh Institute, Sichuan University, Chengdu, 610207, China.}
\thanks{Yuqi Ouyang is with the College of Computer Science, Sichuan University, Chengdu, 610065, China (e-mail: yuqi.ouyang@scu.edu.cn).}
\thanks{Corresponding author: Yuqi Ouyang.}
}

\markboth{Journal of \LaTeX\ Class Files, Vol. 14, No. 8, August 2015}
{Shell \MakeLowercase{\textit{et al.}}: Bare Demo of IEEEtran.cls for IEEE Journals}
\maketitle

\begin{abstract}
Spiking Neural Networks (SNNs) provide energy-efficient computation but their deployment is constrained by dense connectivity and high spiking operation costs. Existing magnitude-based pruning strategies, when naively applied to SNNs, fail to account for temporal accumulation, non-uniform timestep contributions, and membrane stability, often leading to severe performance degradation. This paper proposes Spiking Layer-Adaptive Magnitude-based Pruning (SLAMP), a theory-guided pruning framework that generalizes layer-adaptive magnitude pruning to temporal SNNs by explicitly controlling worst-case output distortion across layers and timesteps. SLAMP formulates sparsity allocation as a temporal distortion–constrained optimization problem, yielding time-aware layer importance scores that reduce to conventional layer-adaptive pruning in single-timestep limit. An efficient two-stage procedure is derived, combining temporal score estimation, global sparsity allocation, and magnitude pruning with retraining for stability recovery. Experiments on CIFAR10, CIFAR100, and the event-based CIFAR10-DVS datasets demonstrate that SLAMP achieves substantial connectivity and spiking operation reductions while preserving accuracy, enabling efficient and deployable SNN inference.
\end{abstract}


\IEEEpeerreviewmaketitle

\section{Introduction}
The human brain achieves remarkable computational efficiency despite operating with billions of synapses \cite{biebl2000analysis, spalding2013dynamics}. Inspired by this property, Spiking Neural Networks (SNNs) emulate cortical computation by transmitting information through discrete spike events rather than continuous activations \cite{maass1997networks, tavanaei2019deep}. This event-driven paradigm offers inherent advantages in energy efficiency, temporal processing, and neuromorphic hardware compatibility. However, modern deep SNNs remain heavily over-parameterized, leading to high spiking operation counts and limiting their practical deployment, particularly under strict energy and latency constraints.

Network pruning is a natural approach to address this challenge \cite{lennie2003cost, neftci2016stochastic,roy2019towards}. In SNNs, sparsity arises from both discrete spiking activity and structural redundancy in synaptic connections. While magnitude-based pruning (MP) has proven effective in artificial neural networks, it typically relies on heuristic layer-wise rules or manual hyperparameter tuning \cite{frankle2018lottery}. Layer-Adaptive Magnitude-based Pruning (LAMP) improves upon this by automatically allocating sparsity across layers to minimize output distortion in static networks \cite{lee2020layer}. However, directly applying such static pruning criteria to SNNs is fundamentally inadequate, as SNN outputs accumulate over time and individual timesteps contribute unevenly to the final decision. Ignoring these temporal effects often leads to membrane instability and severe degradation of temporal fidelity.

Several SNN-specific pruning strategies have attempted to mitigate these issues. Early approaches frequently compromised temporal precision \cite{gale2020sparse, tan2019efficientnet}, while more recent methods exploit activity correlations or surrogate gradient-based regularization to induce dynamic sparsity \cite{zhang2021fast, deng2021comprehensive}. Nevertheless, these methods typically lack an explicit mechanism to control worst-case temporal distortion and often depend on empirical heuristics to balance sparsity and accuracy.

Building on these insights, we propose Spiking Layer-Adaptive Magnitude-based Pruning (SLAMP), a theory-guided pruning framework that generalizes layer-adaptive magnitude pruning to temporal SNNs. SLAMP formulates sparsity allocation as a temporal distortion–constrained optimization problem, explicitly accounting for layer-wise and time-dependent output perturbations induced by pruning. This formulation yields time-aware layer importance scores that automatically adapt sparsity under a global budget and reduce to conventional LAMP in the static or single-timestep limit. An efficient two-stage pruning procedure is derived, combining temporal score estimation, global sparsity allocation, and magnitude pruning with retraining for membrane stability recovery. Our main contributions are:

\begin{itemize}
    \item We propose SLAMP, a distortion-aware pruning framework for SNNs that allocates sparsity by explicitly controlling worst-case temporal output distortion, without heuristic layer exemptions or manual hyperparameter tuning.
    \item We develop a theory-driven extension of layer-adaptive magnitude pruning to temporal SNNs, deriving time-aware layer importance scores that provably generalize static LAMP formulations.
    \item Comprehensive experiments on CIFAR10, CIFAR100, and the event-based CIFAR10-DVS datasets demonstrate that SLAMP substantially reduces connectivity and spiking operations while preserving accuracy and temporal fidelity.
\end{itemize}

\section{Proposed Method}
SLAMP extends layer-adaptive magnitude pruning to spiking neural networks by explicitly accounting for temporal accumulation and uneven timestep contributions. We first briefly introduce the spiking neuron model, then derive a distortion-aware pruning objective for SNNs, and finally describe an efficient two-stage pruning and retraining procedure.

\subsection{Spiking Neural Networks}
We consider SNNs composed of Integrate-and-Fire (IF) neurons. Each neuron integrates incoming spikes from the previous layer while maintaining a membrane potential that resets after firing. At layer $i$ and timestep $t$, the membrane potential update with reset is presented as:
\begin{equation}
H^i_t = H^i_{t-1} \odot (1 - S^i_{t-1}) + (W^i)^\top S^{i-1}_t + H_{\text{reset}} \odot S^i_{t-1},
\end{equation}
and the binary spike generation is:
\begin{equation}
S^i_t = \theta(H^i_t - h),
\end{equation}
where $W^{i}$ is the layer-wise weight matrix, $\theta(\cdot)$ denotes the Heaviside step function. Here, $h$ is the firing threshold determining when the neuron emits a spike, and $H_{\text{reset}}$ is the reset potential to which the membrane returns after firing.

\begin{algorithm}[!t]
\caption{SLAMP for SNNs}
\For{each epoch $e$}{
  \textbf{Fine-Tuned Step}\\
  Simulate IF dynamics over time steps $t=1\ldots T$; \\
  Update weights by surrogate gradient. \\
  \textbf{Pruning Step}\\
  \If{$e \bmod f = 0$}{
    Compute importance score $R^{i}$ in \cref{eq:importance_score}; \\
    Compute pruning mask in \cref{eq:budget_constraint,eq:prun_mask}; \\
    Prune least important weights in \cref{eq:prune}. \\
  }
}
\Return{Pruned and fine-tuned SNN}
\label{algorithm:1}
\end{algorithm}

\subsection{Temporal Distortion–Aware Sparsity Allocation}
Conventional layer-adaptive pruning methods \cite{lee2020layer} are designed for static neural networks and assess weight importance using instantaneous activations. This assumption is violated in spiking neural networks, where synaptic currents accumulate over time through repeated spike integration. As a result, pruning-induced perturbations are inherently temporal, and static criteria may underestimate the distortion caused by removing weights with persistent contributions across timesteps.

We therefore define a temporal importance score that captures the accumulated contribution of each synapse over time. For layer $i$, the importance score is given by:
\begin{equation}
R^{i} = \lambda^{i} \sum_{t=1}^{T}
\big( (W^{i})^\top \mathrm{diag}(S^{i-1}_t) \big)
\odot
\big( (W^{i})^\top \mathrm{diag}(S^{i-1}_t) \big),
\label{eq:importance_score}
\end{equation}
where $\mathrm{diag}(\cdot)$ constructs a diagonal matrix from the presynaptic spike vector, and $\lambda^{i}$ normalizes the scores such that $\sum_j R^{i}_j = 1$. This formulation aggregates squared synaptic contributions across time, assigning higher importance to weights associated with frequently active presynaptic neurons.

\begin{figure}[t]
\centering 
\hspace{-25pt}
\subfigure[Spiking ResNet19]{
\label{Fig.sub.1}
\includegraphics[width = 0.69\linewidth]{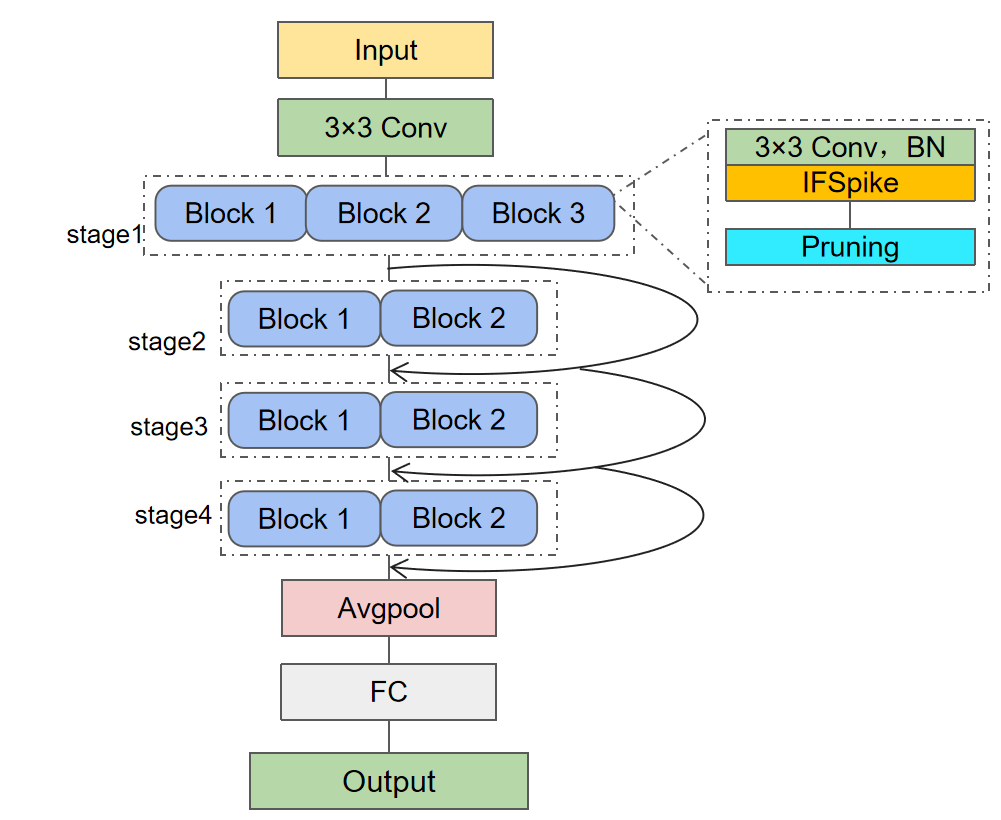}}
\subfigure[Spiking VGGNet]{
\label{Fig.sub.2}
\includegraphics[width = 0.31\linewidth]{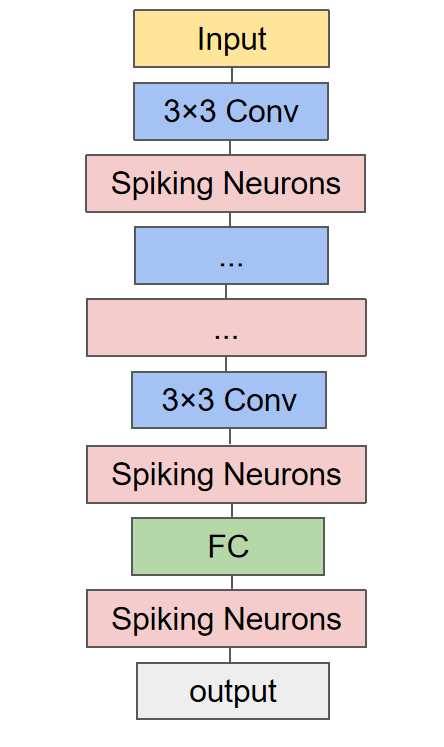}}
\caption{Structures of the two SNNs studied in this work.}
\label{fig:1}
\end{figure}

Pruning is performed under a temporal distortion constraint, defined as:
\begin{equation}
\sum_{j : R^{i}_j < r^{i}} R^{i}_j \le R_{\text{max}},
\label{eq:budget_constraint}
\end{equation}
where $R_{\max}$ specifies the maximum allowable accumulated distortion, $r^{i}$ is a score threshold. Since the importance scores are nonnegative and normalized, this constraint implicitly restricts the size of pruned weights and excludes arbitrarily large pruning sets. In practice, $R_{\max}$ is chosen to be sufficiently small, corresponding to a low-distortion regime.

The induced binary pruning mask and the pruned weight matrix are defined as:
\begin{equation}
M^{i}_j =
\begin{cases}
1, & R^{i}_j \ge r^{i}, \\
0, & R^{i}_j < r^{i},
\end{cases}
\label{eq:prun_mask}
\end{equation}

\begin{equation}
\widetilde{W}^{i} = M^{i} \odot W^{i}.
\label{eq:prune}
\end{equation}

\begin{figure*}[!t]    
\centering
    \includegraphics[width=1\linewidth]{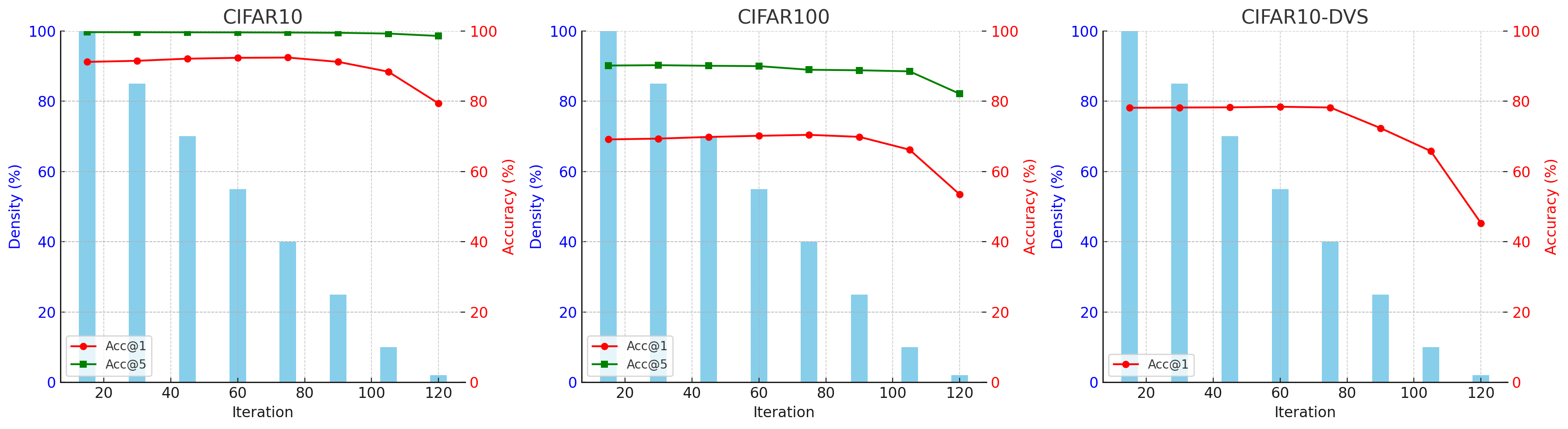}
    \caption{A complete illustration of the pruning performance across the three datasets, including pruned model density (histograms) and resulted Top-1 accuracy (red curve) and Top-5 accuracy (green curve).}
    \label{fig:pruning_process}
\end{figure*}

\begin{table*}[!t]
\centering
\caption{Performance comparisons with previous baseline methods on the three datasets, where Acc. Diff denotes the TOP-1 accuracy difference between pruned and original SNN.}
\label{tab:ours_vs_others}
\resizebox{1\textwidth}{!}{
\begin{tabular}{cllcccccc}
\toprule
\textbf{Dataset} & \textbf{Pruning Method} & \textbf{Architecture} & \textbf{T} & \textbf{Acc. Diff(\%)} & \textbf{Top-1 Acc.(\%)} & \textbf{Conn.(\%)} & \textbf{Param. (M)} & \textbf{SOPs (M)} \\
\midrule
\multirow{6}{*}{CIFAR10}
& ADMM\cite{admm} & 7 Conv, 2 FC & 8 & -0.13 & 90.19 & 25.03 & 15.54 & -- \\
& Grad R\cite{grad_r} & 6 Conv, 2 FC & 8 & -0.30 & 92.54 & 36.72 & 10.43 & -- \\
& ESLSNN\cite{eslsnn} & ResNet19 & 2 & -1.70 & 91.09 & 50.00 & 6.30 & 180.56 \\
& STDS\cite{stds} & 6 Conv, 2 FC & 8 & -0.35 & 92.49 & 11.33 & 1.71 & 147.22 \\
& UPR\cite{shi2024towards} & 6 Conv, 2 FC & 8 & -0.79 & 92.05 & 1.16 & 9.56 & 16.47 \\
& SLAMP (Ours) & ResNet19 & 2 & +1.23 & 92.42 & 40.00 & 4.47 &112.35 \\
\midrule
\multirow{3}{*}{CIFAR100}
& ESLSNN\cite{eslsnn} & ResNet19 & 2 & -0.99 & 73.48 & 50.00 & 6.32 & 186.25 \\
& UPR\cite{shi2024towards} & SEW ResNet18 & 4 & -4.75 & 69.41 & 2.48 & -- & 6.79 \\
& SLAMP (Ours) & ResNet19 & 2 & +0.73 & 69.85 & 25.00 & 4.47 & 115.72 \\
\midrule
\multirow{4}{*}{CIFAR10-DVS}
& ESLSNN\cite{eslsnn} & VGGSNN & 10 & -0.28 & 78.30 & 10.00 & 0.92 & 129.64 \\
& STDS\cite{stds} & VGGSNN & 10 & -2.60 & 79.80 & 4.67 & 0.24 & 38.85 \\
& UPR\cite{shi2024towards} & VGGSNN & 10 & -0.50 & 78.30 & 0.77 & 1.81 & 6.75 \\
& SLAMP (Ours) & VGGSNN & 5 & +0.09 & 78.21 & 40.00 & 3.60 & 105.45 \\
\bottomrule
\end{tabular}
}
\end{table*}

Due to the linear accumulation of synaptic currents in spiking neurons, the output perturbation induced by pruning at layer $i$ can be bounded in terms of the removed synaptic contributions. We assume a low-distortion pruning regime in which the set of pruned weights satisfies the budget constraint in \cref{eq:budget_constraint}. Under this assumption, for any bounded presynaptic spike sequence $\{S^{i-1}_t\}_{t=1}^T$, the temporal deviation of the layer output admits the bound:
\begin{equation}
\sup_{\|S^{i-1}_t\|_\infty \le 1}
\left\|
\sum_{t=1}^{T}
\left(
(W^{i} - \widetilde{W}^{i})^\top S^{i-1}_t
\right)
\right\|_2^2
\;\le\;
C_T \sum_{j : R^{i}_j < r^{i}} R^{i}_j,
\label{eq:temporal_distortion_bound}
\end{equation}
where $C_T$ is a temporal accumulation constant depending only on the spike integration dynamics. Consequently, removing weights in increasing order of $R^{i}$ minimizes an upper bound on the worst-case temporal output distortion subject to the sparsity budget, establishing the optimality of the proposed temporal distortion-aware sparsity allocation rule. In the single-timestep limit ($T=1$) with all presynaptic neurons active, the temporal importance score reduces to the standard LAMP formulation as follows:
\begin{equation}
R^i = \lambda^i \, (W^i)^\top \odot (W^i)^\top.
\end{equation}

We adopt an iterative framework to efficiently realize the proposed temporal distortion–aware pruning objective. As summarized in \cref{algorithm:1}, the SNN is first trained to convergence to establish a stable baseline and reliable temporal activity statistics. Next, synaptic connections are iteratively pruned according to their temporal importance scores, followed by fine-tuning to recover membrane stability and preserve accuracy. This pruning–retraining cycle proceeds progressively until performance degradation occurs, yielding compact SNNs with substantially reduced connectivity and spiking operations while remaining in a low-distortion regime. In practice, the score threshold $r^{i}$ and the distortion constraint $R_{\max}$ are implemented via a predefined pruning schedule that specifies the fraction of weights removed at each step. This schedule serves as a practical surrogate for the distortion-constrained optimization, enabling controlled and stable sparsification across training epochs.

\section{Experiments}

\subsection{Datasets}
We evaluate SLAMP on three datasets that collectively test both static and event-driven spiking networks: CIFAR10, CIFAR100, and CIFAR10-DVS \cite{2009Learning, cheng2020structureawarenetworklanemarker}. CIFAR10 contains 60,000 color images across 10 classes with 50,000 for training and 10,000 for testing, and CIFAR100 includes 100 classes, each containing 600 images with 500 for training and 100 for testing, serving as standard benchmarks for static image classification. CIFAR10-DVS converts 10,000 CIFAR10 images into event streams via a dynamic vision sensor, with 70\% for training and 30\% for evaluation, offering a realistic testbed for event-driven SNNs with temporal dynamics. These datasets allow us to validate SLAMP’s effectiveness across both conventional and temporal spiking settings.

\subsection{Implementation Details}
For CIFAR10 and CIFAR100, we employ Spiking-ResNet19 with IF neurons, convolutional, pooling, and fully-connected layers; CIFAR10-DVS uses Spiking VGGNet with 5-frame input clips, the network architectures are illustrated in \cref{fig:1}. Time steps are set to 2 for static datasets and 5 for CIFAR10-DVS, with learning rates of 0.02 and 0.01, respectively. Networks are initially trained for 300 epochs using SGD with a scheduler. Followed by iteratively pruning $15\%$ of connections and subsequent 15 epochs of fine-tuning. The cycle process continues until connectivity reaches 10\%, after which only 2\% and then 0.4\% connectivity are retained in the final two rounds, yielding highly sparse for experimental study.

\subsection{Performance Comparison}
\cref{fig:pruning_process} illustrates the progressive pruning process. Model density decreases steadily while Top-1 and Top-5 accuracies remain stable until extreme sparsity triggers a sharp drop. This demonstrates that SLAMP can remove a substantial portion of parameters while maintaining both classification performance and temporal consistency in spiking activity.

\cref{tab:ours_vs_others} compares SLAMP with baseline pruning methods. On CIFAR10, SLAMP achieves $+1.23\%$ Top-1 improvement at $40\%$ connectivity, outperforming alternatives that either degrade accuracy or maintain higher complexity. For CIFAR100, SLAMP maintains a positive accuracy difference even under low connectivity, while competing methods suffer notable performance loss. On CIFAR10-DVS, SLAMP preserves event-driven classification accuracy with 40\% connectivity, highlighting the framework’s capability to retain temporal fidelity under aggressive pruning. These results confirm that SLAMP achieves a favorable accuracy–efficiency trade-off and preserves membrane dynamics, supporting its deployment for energy-constrained SNNs.

\subsection{Ablation Study of Fine-Tuning}
We evaluate the effect of fine-tuning after pruning on network stability and performance. Membrane variance, defined as the variance of membrane potential across time averaged over neurons, serves as a proxy for temporal fidelity and spiking sparsity. \cref{tab:ablation_full} shows that fine-tuning substantially improves accuracy, particularly at low connectivity, while significantly reducing membrane variance. This confirms that fine-tuning is essential for restoring temporal expressiveness and maintaining stable spiking dynamics, a key requirement for the theoretical guarantees of SLAMP.

\begin{table}[!t]
\centering
\caption{Accuracy and membrane potential variance with and without fine-tuning, across different datasets and connectivity.}
\label{tab:ablation_full}
\resizebox{0.48\textwidth}{!}{
\begin{tabular}{c c c c c c}
\toprule
\multirow{2}{*}{Dataset} & \multirow{2}{*}{Connectivity} & \multicolumn{2}{c}{Accuracy (\%)} & \multicolumn{2}{c}{Membrane Variance} \\
\cmidrule(lr){3-4} \cmidrule(lr){5-6}
 & & With & Without & With & Without \\
\midrule
\multirow{4}{*}{CIFAR10} & 70\% & 92.10 & 88.32 & 0.081 & 0.153 \\
 & 40\% & \textbf{92.42} & 84.17 & 0.092 & 0.221 \\
 & 25\% & 91.20 & 76.45 & 0.105 & 0.298 \\
 & 10\% & 88.40 & 63.28 & 0.134 & 0.412 \\
\midrule
\multirow{4}{*}{CIFAR100} & 70\% & 69.82 & 65.14 & 0.095 & 0.182 \\
 & 40\% & \textbf{70.42} & 61.87 & 0.103 & 0.254 \\
 & 25\% & 69.85 & 55.33 & 0.118 & 0.331 \\
 & 10\% & 66.20 & 46.82 & 0.152 & 0.447 \\
\midrule
\multirow{4}{*}{CIFAR10-DVS} & 70\% & \textbf{78.25} & 74.61 & 0.112 & 0.201 \\
 & 40\% & 78.21 & 70.15 & 0.125 & 0.278 \\
 & 25\% & 72.34 & 63.40 & 0.141 & 0.359 \\
 & 10\% & 65.80 & 52.17 & 0.178 & 0.483 \\
\bottomrule
\end{tabular}
}
\end{table}

\begin{table}[!t]
\centering
\caption{Hyperparameter analysis of pruning frequency and learning rate across datasets.}
\label{tab:sensitivity}
\scalebox{1}{
\resizebox{0.48\textwidth}{!}{
\begin{tabular}{c c c c c c c}
\toprule
\multirow{2}{*}{Hyperparameter} & \multirow{2}{*}{Value} & \multicolumn{2}{c}{CIFAR10} & \multicolumn{2}{c}{CIFAR10-DVS} & \multirow{2}{*}{Sparsity} \\
\cmidrule(lr){3-4} \cmidrule(lr){5-6}
& & Accuracy & $\Delta$Acc & Accuracy & $\Delta$Acc & Deviation \\
\midrule
\multirow{3}{*}{Pruning Freq $f$} 
& 10 & 92.15\% & -0.27 & 77.92\% & -0.29 & 1.8\% \\
& 15 & \textbf{92.42\%} & 0.00 & \textbf{78.21\%} & 0.00 & 0.9\% \\
& 25 & 92.38\% & -0.04 & 78.15\% & -0.06 & 1.2\% \\
\midrule
\multirow{3}{*}{Learning Rate $\eta_0$} 
& 0.005 & 91.25\% & -1.17 & 76.84\% & -1.37 & 2.1\% \\
& 0.01 & 92.03\% & -0.39 & \textbf{78.21\%} & 0.00 & 1.3\% \\
& 0.02 & \textbf{92.42\%} & 0.00 & 77.92\% & -0.29 & 0.9\% \\
\bottomrule
\end{tabular}
}
}
\end{table}

\subsection{Hyperparameter Analysis}
We evaluate the robustness of SLAMP with respect to two key hyperparameters: the pruning frequency $f$, defined as the number of epochs between successive pruning steps, and the initial learning rate $\eta_0$. As reported in \cref{tab:sensitivity}, SLAMP exhibits stable Top-1 accuracy and consistently small performance drops ($\Delta$Acc), while maintaining sparsity deviations below a narrow margin across all tested settings. Furthermore, optimal performance is achieved at $(f,\eta_0)=(15,0.02)$ for CIFAR10 and $(15,0.01)$ for CIFAR10-DVS. These results supports our hyperparameter selection, also showing that SLAMP maintains both temporal fidelity and accuracy across practical hyperparameter ranges, thus reducing the need for extensive tuning and demonstrating suitability for real-world SNN deployment.

\section{Conclusion}
We introduced SLAMP, a theory-driven pruning framework for Spiking Neural Networks that extends LAMP to account for layer-wise and temporal perturbations. By formulating pruning as a temporal distortion–minimization problem, SLAMP automatically identifies and removes less important weights while preserving membrane dynamics and temporal fidelity. SLAMP first stabilizes network parameters and then iteratively prunes and fine-tunes the network. Experiments on CIFAR10, CIFAR100, and CIFAR10-DVS demonstrate that SLAMP achieves high sparsity, reduces spiking operations, and maintains or even improves predictive accuracy. These findings highlight SLAMP as an effective approach for deploying energy-efficient, temporally consistent SNNs in resource-constrained environments.

\vfill\pagebreak

\bibliographystyle{IEEEtran}
\small
\bibliography{refs}

@article{biebl2000analysis,
  title={Analysis of neurogenesis and programmed cell death reveals a self-renewing capacity in the adult rat brain},
  author={Biebl, Manfred and Cooper, Christiana M and Winkler, J{\"u}rgen and Kuhn, H Georg},
  journal={Neuroscience letters},
  volume={291},
  number={1},
  pages={17--20},
  year={2000},
  publisher={Elsevier}
}

@article{spalding2013dynamics,
  title={Dynamics of hippocampal neurogenesis in adult humans},
  author={Spalding, Kirsty L and Bergmann, Olaf and Alkass, Kanar and Bernard, Samuel and Salehpour, Mehran and Huttner, Hagen B and Bostr{\"o}m, Emil and Westerlund, Isabelle and Vial, C{\'e}line and Buchholz, Bruce A and others},
  journal={Cell},
  volume={153},
  number={6},
  pages={1219--1227},
  year={2013},
  publisher={Elsevier}
}

@article{maass1997networks,
  title={Networks of spiking neurons: the third generation of neural network models},
  author={Maass, Wolfgang},
  journal={Neural networks},
  volume={10},
  number={9},
  pages={1659--1671},
  year={1997},
  publisher={Elsevier}
}

@article{neftci2016stochastic,
  title={Stochastic synapses enable efficient brain-inspired learning machines},
  author={Neftci, Emre O and Pedroni, Bruno U and Joshi, Siddharth and Al-Shedivat, Maruan and Cauwenberghs, Gert},
  journal={Frontiers in neuroscience},
  volume={10},
  pages={241},
  year={2016},
  publisher={Frontiers Media SA}
}

@article{lennie2003cost,
  title={The cost of cortical computation},
  author={Lennie, Peter},
  journal={Current biology},
  volume={13},
  number={6},
  pages={493--497},
  year={2003},
  publisher={Elsevier}
}

@article{tavanaei2019deep,
  title={Deep learning in spiking neural networks},
  author={Tavanaei, Amirhossein and Ghodrati, Masoud and Kheradpisheh, Saeed Reza and Masquelier, Timoth{\'e}e and Maida, Anthony},
  journal={Neural networks},
  volume={111},
  pages={47--63},
  year={2019},
  publisher={Elsevier}
}

@article{frankle2018lottery,
  title={The lottery ticket hypothesis: Finding sparse, trainable neural networks},
  author={Frankle, Jonathan and Carbin, Michael},
  journal={arXiv preprint arXiv:1803.03635},
  year={2018}
}

@article{lee2020layer,
  title={Layer-adaptive sparsity for the magnitude-based pruning},
  author={Lee, Jaeho and Park, Sejun and Mo, Sangwoo and Ahn, Sungsoo and Shin, Jinwoo},
  journal={arXiv preprint arXiv:2010.07611},
  year={2020}
}

@article{roy2019towards,
  title={Towards spike-based machine intelligence with neuromorphic computing},
  author={Roy, Kaushik and Jaiswal, Akhilesh and Panda, Priyadarshini},
  journal={Nature},
  volume={575},
  number={7784},
  pages={607--617},
  year={2019},
  publisher={Nature Publishing Group UK London}
}

@inproceedings{gale2020sparse,
  title={Sparse gpu kernels for deep learning},
  author={Gale, Trevor and Zaharia, Matei and Young, Cliff and Elsen, Erich},
  booktitle={SC20: International \ Conference for High Performance Computing, Networking, Storage and Analysis}, 
  pages={1--14}, 
  year={2020},
  organization={IEEE}
}

@inproceedings{tan2019efficientnet,
  title={Efficientnet: Rethinking model scaling for convolutional neural networks},
  author={Tan, Mingxing and Le, Quoc},
  booktitle={International conference on machine learning},
  pages={6105--6114},
  year={2019},
  organization={PMLR}
}

@article{zhang2021fast,
  title={A fast spiking neural network accelerator based on BP-STDP algorithm and weighted neuron model},
  author={Zhang, Jian and Wang, Ran and Pei, Xudong and Luo, Dan and Hussain, Sajjad and Zhang, Guohe},
  journal={IEEE Transactions on Circuits and Systems II: Express Briefs},
  volume={69},
  number={4},
  pages={2271--2275},
  year={2021},
  publisher={IEEE}
}

@article{deng2021comprehensive,
  title={Comprehensive snn compression using admm optimization and activity regularization},
  author={Deng, Lei and Wu, Yujie and Hu, Yifan and Liang, Ling and Li, Guoqi and Hu, Xing and Ding, Yufei and Li, Peng and Xie, Yuan},
  journal={IEEE transactions on neural networks and learning systems},
  volume={34},
  number={6},
  pages={2791--2805},
  year={2021},
  publisher={IEEE}
}

@article{2009Learning,
  title={Learning multiple layers of features from tiny images},
  author={ Krizhevsky, A.  and  Hinton, G. },
  journal={Handbook of Systemic Autoimmune Diseases},
  volume={1},
  number={4},
  year={2009},
}

@misc{cheng2020structureawarenetworklanemarker,
      title={Structure-Aware Network for Lane Marker Extraction with Dynamic Vision Sensor}, 
      author={Wensheng Cheng and Hao Luo and Wen Yang and Lei Yu and Wei Li},
      year={2020},
      eprint={2008.06204},
      archivePrefix={arXiv},
      primaryClass={cs.CV},
      url={https://arxiv.org/abs/2008.06204}, 
}

@misc{admm,
      title={Comprehensive SNN Compression Using ADMM Optimization and Activity Regularization}, 
      author={Lei Deng and Yujie Wu and Yifan Hu and Ling Liang and Guoqi Li and Xing Hu and Yufei Ding and Peng Li and Yuan Xie},
      year={2020},
      eprint={1911.00822},
      archivePrefix={arXiv},
      primaryClass={cs.NE},
      url={https://arxiv.org/abs/1911.00822}, 
}

@inproceedings{grad_r, series={IJCAI-2021},
   title={Pruning of Deep Spiking Neural Networks through Gradient Rewiring},
   url={http://dx.doi.org/10.24963/ijcai.2021/236},
   DOI={10.24963/ijcai.2021/236},
   booktitle={Proceedings of the Thirtieth International Joint Conference on Artificial Intelligence},
   publisher={International Joint Conferences on Artificial Intelligence Organization},
   author={Chen, Yanqi and Yu, Zhaofei and Fang, Wei and Huang, Tiejun and Tian, Yonghong},
   year={2021},
   month=aug, pages={1713–1721},
   collection={IJCAI-2021} }

@misc{eslsnn,
      title={Improving the Sparse Structure Learning of Spiking Neural Networks from the View of Compression Efficiency}, 
      author={Jiangrong Shen and Qi Xu and Gang Pan and Badong Chen},
      year={2025},
      eprint={2502.13572},
      archivePrefix={arXiv},
      primaryClass={cs.HC},
      url={https://arxiv.org/abs/2502.13572}, 
}

@InProceedings{stds,
  title = 	 {State Transition of Dendritic Spines Improves Learning of Sparse Spiking Neural Networks},
  author =       {Chen, Yanqi and Yu, Zhaofei and Fang, Wei and Ma, Zhengyu and Huang, Tiejun and Tian, Yonghong},
  booktitle = 	 {Proceedings of the 39th International Conference on Machine Learning},
  pages = 	 {3701--3715},
  year = 	 {2022},
  editor = 	 {Chaudhuri, Kamalika and Jegelka, Stefanie and Song, Le and Szepesvari, Csaba and Niu, Gang and Sabato, Sivan},
  volume = 	 {162},
  series = 	 {Proceedings of Machine Learning Research},
  month = 	 {17--23 Jul},
  publisher =    {PMLR},
  url = 	 {https://proceedings.mlr.press/v162/chen22ac.html}
}

@inproceedings{shi2024towards,
  title={Towards energy efficient spiking neural networks: An unstructured pruning framework},
  author={Shi, Xinyu and Ding, Jianhao and Hao, Zecheng and Yu, Zhaofei},
  booktitle={The Twelfth International Conference on Learning Representations},
  year={2024}
}

\end{document}